\documentclass{article}
    \PassOptionsToPackage{square,sort,comma,numbers}{natbib}

\usepackage[preprint]{neurips_2020}

\usepackage[utf8]{inputenc} 
\usepackage[T1]{fontenc}    
\usepackage{hyperref}       
\usepackage{url}            
\usepackage{booktabs}       
\usepackage{amsfonts}       
\usepackage{nicefrac}       
\usepackage{microtype}      
\usepackage{wrapfig}
\usepackage{graphicx}
\usepackage{subfigure}
\usepackage{tikz}
\usepackage{xcolor}
\usepackage{multirow}
\usepackage{multicol}

\title{\emph{SeqXFilter}: A Memory-efficient Denoising Filter for Dynamic Vision Sensors}

\author{%
  Shasha Guo, Lei Wang, Xiaofan Chen, Limeng Zhang, Ziyang Kang, Weixia Xu \\
  College of Computer Science and Technology\\
  National University of Defense Technology\\
  Changsha, China \\
  \texttt{guoshasha13@nudt.edu.cn} \\
}

\begin{document}

\maketitle

\begin{abstract}
Neuromorphic event-based dynamic vision sensors (DVS) have much faster sampling rates and a higher dynamic range than frame-based imaging sensors. However, they are sensitive to background activity (BA) events that are unwanted. There are some filters for tackling this problem based on spatio-temporal correlation. However, they are either memory-intensive or computing-intensive. We propose \emph{SeqXFilter}, a spatio-temporal correlation filter with only a past event window that has an O(1) space complexity and has simple computations. We explore the spatial correlation of an event with its past few events by analyzing the distribution of the events when applying different functions on the spatial distances. We find the best function to check the spatio-temporal correlation for an event for \emph{SeqXFilter}, best separating real events and noise events. We not only give the visual denoising effect of the filter but also use two metrics for quantitatively analyzing the filter's performance. Four neuromorphic event-based datasets, recorded from four DVS with different output sizes, are used for validation of our method. The experimental results show that \emph{SeqXFilter} achieves similar performance as baseline NNb filters, but with extremely small memory cost and simple computation logic.
\end{abstract}

\section{Introduction}
Research on neuromorphic event-based sensors (``silicon retinae'') started a few decades back \cite{mead91}. Recently, the technology has matured to a stage where there have been some commercially available sensors. Some of the popular sensors are Dynamic Vision Sensor (DVS) \cite{dvs128}, Asynchronous Time-based Image Sensor (ATIS) \cite{atis}, the sensitive DVS \cite{sDVS}, the Dynamic and Active pixel Vision Sensor (DAVIS) \cite{davis} and the Celex-IV \cite{bib:CeleX}.
Different from conventional frame-based imaging sensors that work by sampling the scene at a fixed temporal rate (typically 30 frames per second \cite{kameyama2016demonstration}), these neuromorphic sensors detect dynamic changes in illumination. This results in a higher dynamic range, higher sampling rate, and lower power consumption. They have been utilized in applications such as pose estimation \cite{poseestimation}, gesture-based remote control \cite{remotecontrol}, corner detection \cite{cornerdetection}, and drones \cite{mueggler2014event}.

However, these sensors will produce background activity (BA) events under constant illumination, which are caused by temporal noise and junction leakage currents \cite{dvs128,Bs2filter,phdthesis}.
There are already multiple noise filtering methods for event-based data available.

The most commonly employed method is the Nearest Neighbor (NNb) filter based on spatio-temporal correlation \cite{Bs2filter, Bs1filter, Bs3filter}. 
However, these spatio-temporal NNb filters are memory intensive. For a DVS with N$\times$N pixels, the minimum space complexity is at O(N) level, which puts a burden on the limited programmable logic (PL) embedded in the sensor head. There are two main reasons for the high space complexity of the current spatio-temporal filters. First, the memory size of these filters are dependent on the DVS output size $N$ for the convenience of detecting spatio-temporal correlation. Second, memory cells of these filters are designed to store the timestamp and possible more information, i.e., at least 32-bits per cell. These two features limit the possibility of further optimizing the memory.

There are also some filters utilizing spatio-temporal correlation \cite{TNfilter, feng2020event, guo2020hashheat} but not storing the timestamp. \cite{TNfilter} uses spiking neurons to reprocess the event. \cite{feng2020event} calculate a density matrix for each event based on its spatio-temporal neighborhood. \cite{guo2020hashheat} uses some hashing functions to encode the spatio-temporal information and stores access values for checking. These methods eliminates the need for storing timestamps but still requires other memory cost and complex computation cost compared with the above NNb filters.

To tackle the memory challenges, we consider the spatio-temporal correlation from another perspective. The previous NNb filters consider storing the spatio-temporal correlation by storing the timestamp for each pixel. The location information of the event, as well as the constraint for spatial correlation, is provided by the memory cell for each pixel or a group of pixels. So the opposite solution is to store the location of few events before the current event for checking spatio-temporal correlation and make the time information as implicit information indicating by the event order in an event stream. The advantage of this solution is that it eliminates the need for assigning huge memory cells for each pixel or a small group of pixels, instead of just assigning memory cells for the location coordinates of the past few events. And it doesn't need to reassign such location memory cells for new coming events, just reusing the memory for every event by rewriting the location of the new event in the memory. Thus, to check the difference between timestamp turns to check the spatial difference between the current event and the past few events. Thus, we propose to maintain a past event window for each event and update the window for each event. That is, the only memory cost is the cost for the past event window. And the window stores the location coordinates of the past events rather than the timestamps of the past events.

There are two challenges to this method. One is how many past events are appropriate to store. Obviously the less the better for memory reduction. The other is how to check the correlation with the past event window. If storing more than one past event, the spatial correlation between these events and the current event could be very different since there could be a noise event in the middle of two real events in the event output sequence. 
We will tackle these challenges in this work.

Our contributions are as follows. First, We propose a memory-efficient filter \emph{SeqXFilter}, checking the spatio-temporal correlation with only a past event window. It has an O(1) space complexity and simple computations. Second, we explore the spatial correlation between an event and its past few events by analyzing the distribution of the events when applying different functions on the spatial distances. We find the function for best separating real events and noise events. Third, we not only give the visual denoising effect of the filter but also use two metrics, PSNR and SSIM, for quantitatively analyzing the filter's performance. Four neuromorphic event-based datasets, recorded from four DVS with different output sizes, are used for validation of our method. The experimental results show that \emph{SeqXFilter} achieves similar performance as baseline NNb filters, but with extremely small memory cost and simple computation logic.

\subsection{DVS}
\label{sec:CeleX}
The DVS128 \cite{dvs128} sensor is an event-based image sensor that generates asynchronous events when it detects the changes in log intensity. If the change of illumination exceeds an upper or lower threshold, the DVS128 will generate an "ON" event or "OFF" event respectively.
DAVIS \cite{davis} combines the DVS with an active pixel sensor (APS) at the pixel level. It allows simultaneous output of asynchronous events and synchronous frames.
The CeleX-IV is a high resolution dynamic vision sensor from CelePixel Technology Co., Ltd. \cite{bib:CeleX}. The resolution of the sensor is 768$\times$640 that is much larger than the maximum output of DAVIS (346$\times$260).

To encode all the event information for output, the DVS128 and DAVIS use the Address Event Representation (AER) protocol \cite{aerprotocal} to create a quadruplets $e(p,x,y,ts)$ for each event. Specifically, $p$ is polarity, i.e., ON or OFF, x is the x-position of a pixel's event, y is the y-position of a pixel's event, and $ts$ is a 32-bits timestamp, i.e., the timing information of an event. The output of CeleX is different \cite{bib:CeleX}. We parse and reconstruct it to be representations like AER.

\subsection{Spatio-temporal correlation}
\begin{wrapfigure}{l}{0.35\textwidth}
        \includegraphics[width=0.2\textheight]{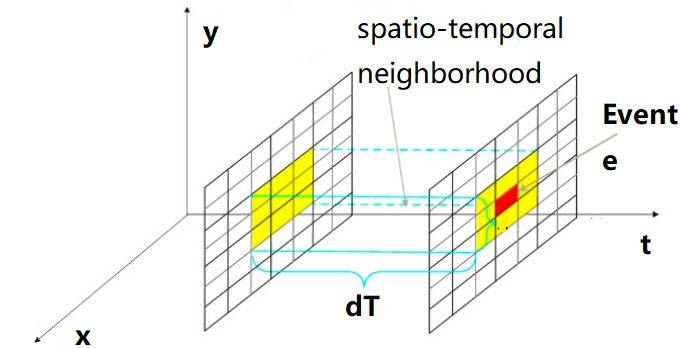}
        \caption{Illustration of the spatio-temporal correlation. The red pixel indicates the location of the current event $e$. The yellow pixels show its spatial neighborhood, and the $dT$ in t-axis suggests the temporal neighborhood.}
\label{fig:spatio-temporal}
\vspace{-0.02in}
\end{wrapfigure}
BA events are caused by thermal noise and junction leakage currents \cite{dvs128,Bs2filter,phdthesis}. These events degrade the quality of the data and further incurs unnecessary communication bandwidth and computing resources.
The BA and the real activity events differ in that the BA event lacks temporal correlation with events in its spatial neighborhood while the real events, arising from moving objects or changes in illumination, have a temporal correlation with events from their spatial neighbors.
Based on this difference, the BA events can be filtered out by detecting events which do not have a spatial correlation with events generated by the neighborhood pixels. Such a filter is a spatio-temporal correlation filter. An example of the spatio-temporal neighborhood for each event is shown in Figure~\ref{fig:spatio-temporal}. 
Any event occurred in the spatial neighborhood within the temporal neighborhood is regarded as having spatio-temporal correlation with the event $e$. The condition can be formulated as the equation $T_{NNb} - T_{e} < dT$. If the condition is met, the event is regarded as a real activity event. The $T_{NNb}$ is the timestamps from the neighborhood pixels, which meet this condition: $|x_{p} - x| \leq 1$ and $|y_{p} - y| \leq 1$ where $p$ stands for a pixel. And $dT$ is the limitation for timestamp difference.


\subsection{Related work}
\label{sec:relatedwork}
\begin{wrapfigure}{l}{0.36\textwidth}
        \subfigure[Bs1]{
  \label{fig:bs1} 
  \includegraphics[width=0.17\textwidth]{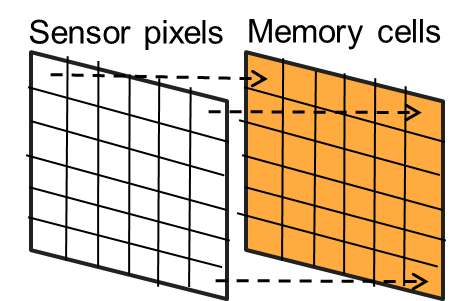}}
  \subfigure[Bs2]{
  \label{fig:bs2} 
  \includegraphics[width=0.17\textwidth]{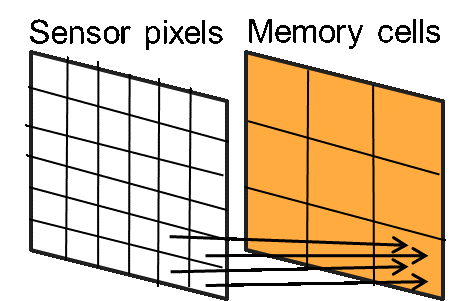}}
\caption{Example of Memory-intensive filters. In bs1, each pixel has a memory cell. In bs2, a group of pixels ($s\times s$) shares a memory cell. A memory cell is used for storing a 32-bits timestamp.}
\label{fig:filters}
\vspace{-0.02in}
\end{wrapfigure}
Here we introduce several filters based on spatio-temporal correlation by storing timestamps or by doing some preprocessing on the spatio-temporal information x, y, and ts.

Delbruck \cite{Bs1filter} proposed a filter where each pixel has a memory cell for storing the last event’s timestamp. The stored timestamps are used for computing the spatio-temporal correlation (Figure~\ref{fig:bs1}). \textbf{We denote this as bs1.}
Liu et al. \cite{Bs2filter} uses sub-sampling groups to reduce the memory size. Each sub-sampling group of factor $s$ includes $s^{2}$ pixels and uses one memory cell for storing the timestamp of the most recent event of the group (Figure~\ref{fig:bs2}). \textbf{We denote this as bs2}.
Khodamoradi et al. \cite{Bs3filter} assigns two memory cells to each row and each column, to store the most recent event in that row or column. 
Both the two cells are 32-bits with one for storing the timestamp and one for polarity and the other axis position.
Padala et al. \cite{TNfilter} proposed a filter with neuromorphic integrate-and-fire neurons which integrate spikes not only from the corresponding pixel but also its neighborhood pixels for firing, each neuron corresponding to a pixel.
Feng et al. \cite{feng2020event} proposed to
calculate the density of an event based on the density matrix maintained for each pixel for deciding the event to be real or noise.
Guo et al. \cite{guo2020hashheat} proposed to use hashing functions to encode the spatio-temporal information of an event and used a list structure to store the encoded correlation information for checking the event.
These filters are either memory-intensive or computing-intensive.
\section{Methodology}
We attempt to propose a filter that checks the spatio-temporal correlation only relying on the past few events before the event, rather than based on its located pixel. We name the filter as SeqXFilter.

We define a past event window with length $X$ to refer to the past few events before the current event. We calculate the spatial distance between the current event and the events in the past window. The \textbf{spatial distance} between the current event ($e_{n}$) and the event in the past window is denoted as $D_{n, n-j}$, where $j$ is the order of the event in the past window counting from the end near the current event. The distance is normalized, i.e., $D_{n, n-j}$ = $|x_{n} - x_{n-j}| / M + |y_{n} - y_{n-j}|/N$, where $j\in \left \{ 1,2,\cdots, X \right \}$, M and N are the width and height of the DVS output. For a past window with length $X$, there are $X$ spatial distances for each event except the first $X$ events in the DVS output.
We do not need the timestamp information explicitly as it already implies in the event sequence.

The length of the past event window will be larger than 1 for the following consideration. When two consecutive real events with spatio-temporal correlation are separated by a noise event in the output stream in the time axis, the latter event is expected to be still supported by the real event before the noise event, to be judged as a real event. If the window is 1, the latter real event is likely to be misjudged since it is almost unlikely to get a spatio-temporal correlation from a noise event.

For the random length of the past event window ($X\textgreater 1$), it remains to be explored how to utilize these spatial distances for comprehensively evaluating the spatio-temporal correlation between the current event and these past events. We introduce a function as $f(D_{n,n-1}, D_{n,n-2}, \cdots, D_{n,n-X})$ to connect these features for checking the event. There are several functions $f$ for connecting these features, such as maximum, $max(D_{n,n-1}, D_{n,n-2}, \cdots, D_{n,n-X})$, minimum, $min(D_{n,n-1}, D_{n,n-2}, \cdots, D_{n,n-X})$, average, $(D_{n,n-1} + D_{n,n-2} + \cdots + D_{n,n-X}) / X$ and weighted average, $(D_{n,n-1} \times a_{1} + D_{n,n-2} \times a_{2} + \cdots + D_{n,n-X} \times a_{X}) / X$. Maximum represents the distance between the current event and the event farthest from the current event in the past event window. Minimum represents the distance between the current event and the event closest to the current event in the past event window. Average represents the average distance between events in the past event window and the current event. Weighted average stands for the average of the weighted distance between events in the past event window and the current event. For example, by giving large weight to the event with minimum distance to the current event and small weight to the event with maximum distance to the current event, the hypothesis is that events that are closer to the current event are more important for measuring the spatio-temporal correlation.

When we decide the best function, which chooses a value from the spatial distances as spatial correlation, we also need a threshold $\sigma$. When the value for an event is smaller than $\sigma$, it is regarded as a real event, otherwise, noise event.

\begin{figure}
\centering
\subfigure[When event $n$ comes]{
   \label{fig:ov1} 
   \includegraphics[width=0.3\textwidth]{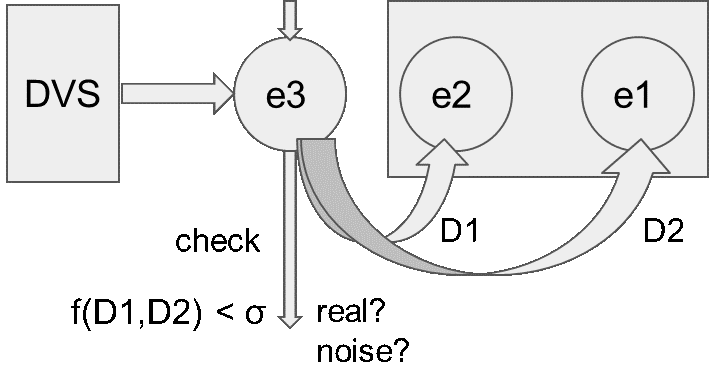}}
   \hspace{0.2in}
   \subfigure[When event $n+1$ comes]{
   \label{fig:ov2} 
   \includegraphics[width=0.3\textwidth]{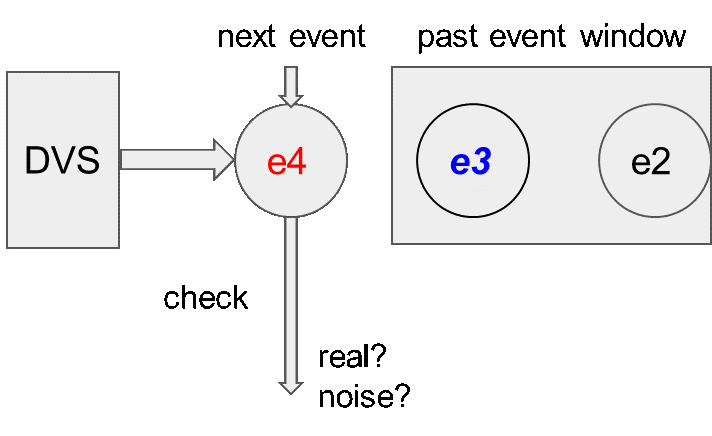}}
\caption{Illustration of the scheme proposed in this paper. For each event, it is checked as Figure~\ref{fig:ov1} shows. After checking, the event is stored in the past event window (x and y coordinates) for the next event's checking as Figure~\ref{fig:ov2} shows.}
\label{fig:overview}
\end{figure}

The illustration of the working scheme with an example of $X=2$ is shown in Figure~\ref{fig:overview}.

\begin{figure}[htbp]
\subfigure[avg]{
    \begin{minipage}[t]{0.24\textwidth}
        \centering
        \includegraphics[width=1.35in, height=0.8in]{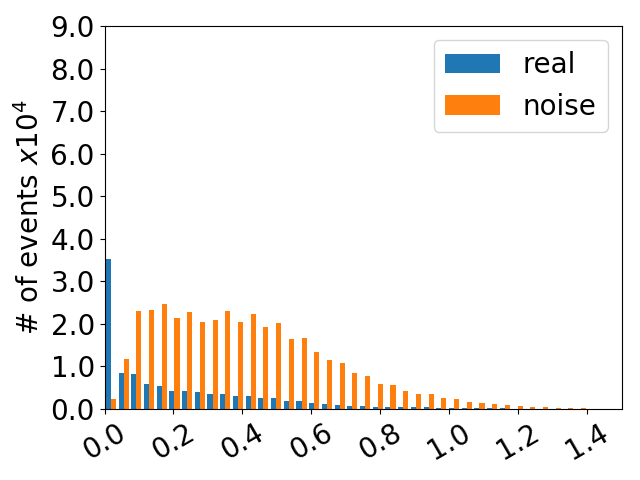}\\
        \vspace{0.02cm}
        \includegraphics[width=1.35in, height=0.8in]{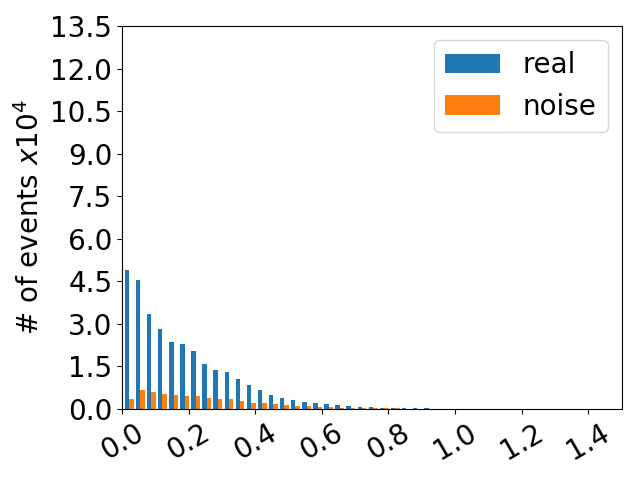}\\
        \vspace{0.02cm}
        \includegraphics[width=1.35in, height=0.8in]{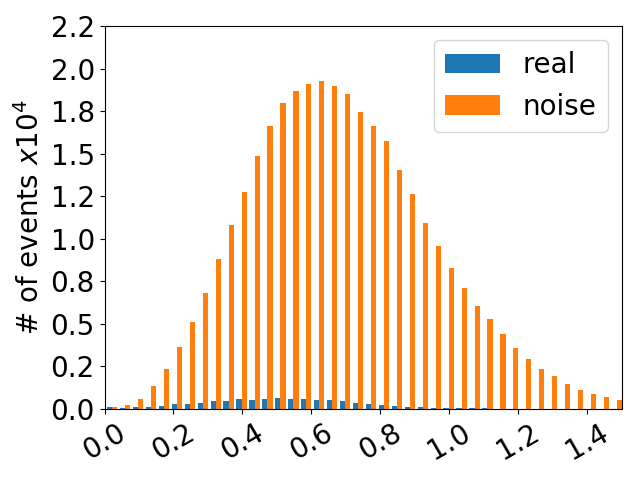}\\
        \vspace{0.02cm}
        \includegraphics[width=1.35in, height=0.8in]{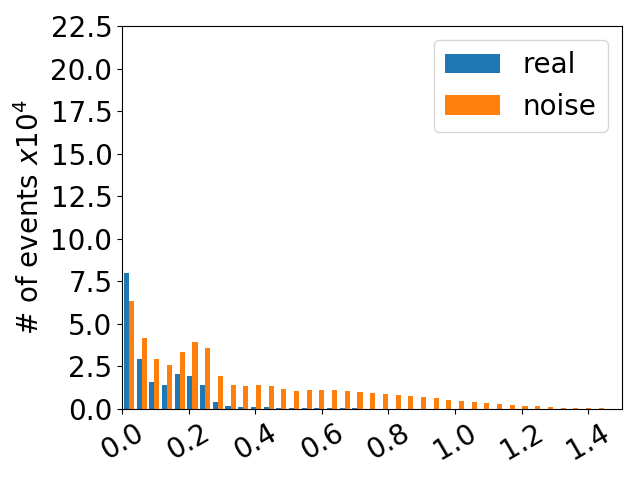}\\
        \vspace{0.02cm}
    \end{minipage}%
}%
\subfigure[avg31]{
    \begin{minipage}[t]{0.24\textwidth}
        \centering
        \includegraphics[width=1.35in, height=0.8in]{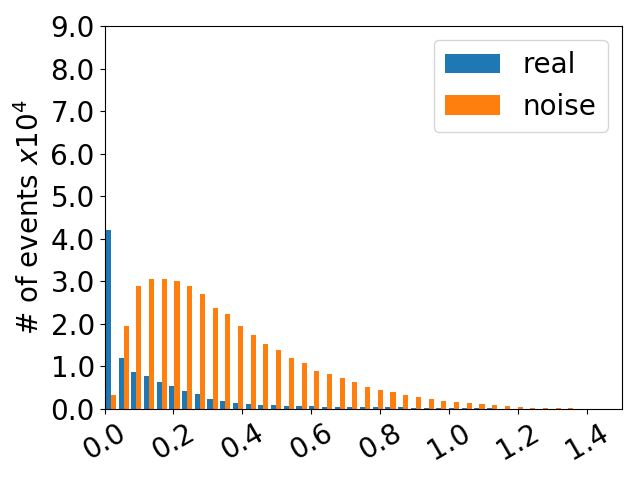}\\
        \vspace{0.02cm}
        \includegraphics[width=1.35in, height=0.8in]{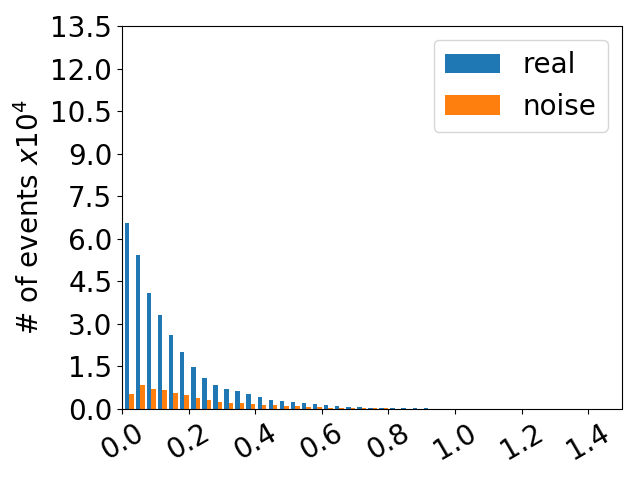}\\
        \vspace{0.02cm}
        \includegraphics[width=1.35in, height=0.8in]{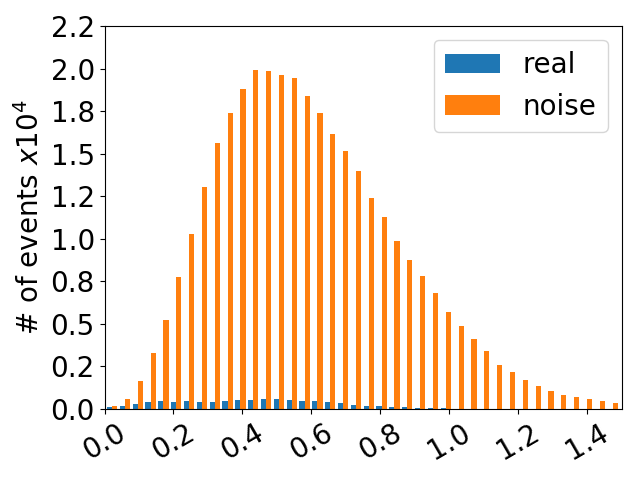}\\
        \vspace{0.02cm}
        \includegraphics[width=1.35in, height=0.8in]{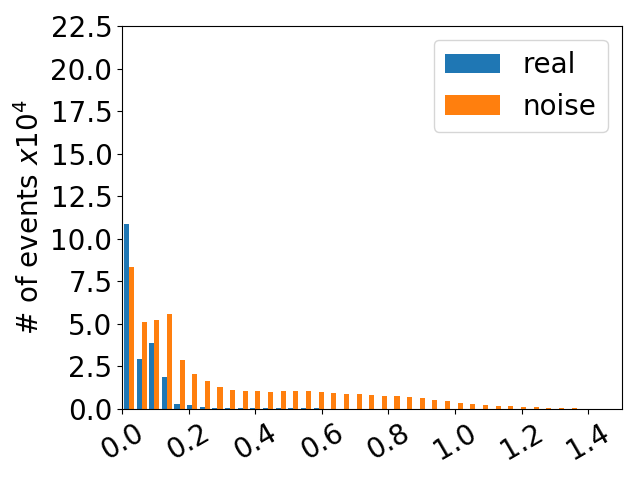}\\
        \vspace{0.02cm}
    \end{minipage}%
}%
\subfigure[min]{
    \begin{minipage}[t]{0.24\textwidth}
        \centering
        \includegraphics[width=1.35in, height=0.8in]{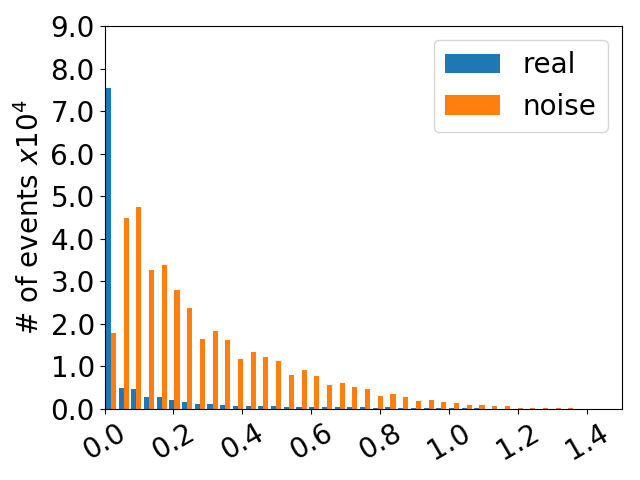}\\
        \vspace{0.02cm}
        \includegraphics[width=1.35in, height=0.8in]{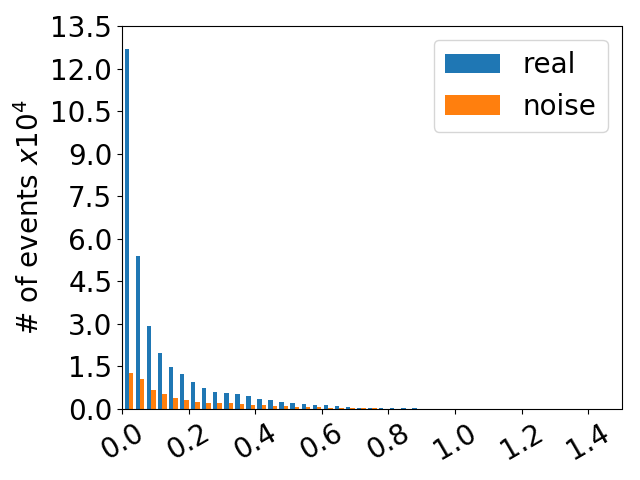}\\
        \vspace{0.02cm}
        \includegraphics[width=1.35in, height=0.8in]{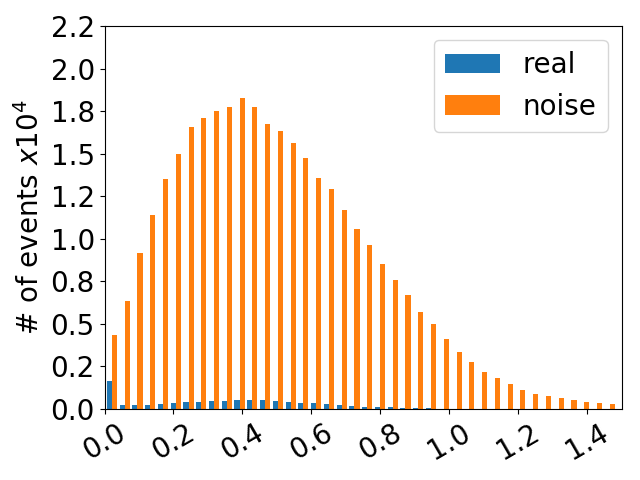}\\
        \vspace{0.02cm}
        \includegraphics[width=1.35in, height=0.8in]{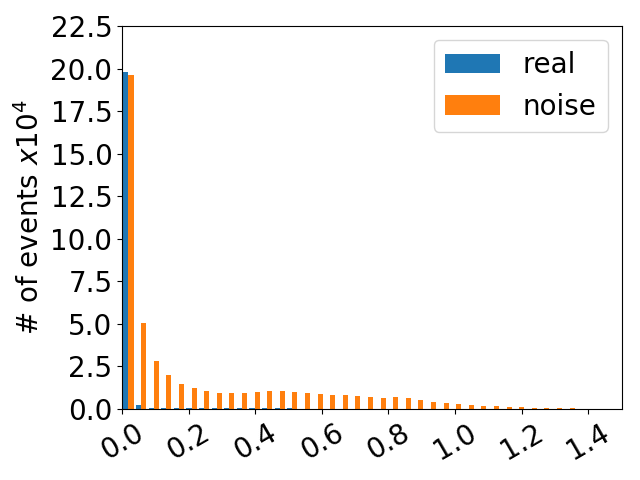}\\
        \vspace{0.02cm}
    \end{minipage}%
}%
\subfigure[max]{
    \begin{minipage}[t]{0.24\textwidth}
        \centering
        \includegraphics[width=1.35in, height=0.8in]{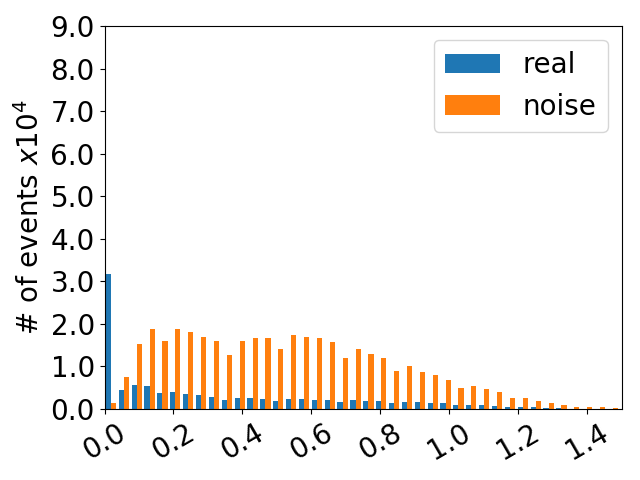}\\
        \vspace{0.02cm}
        \includegraphics[width=1.35in, height=0.8in]{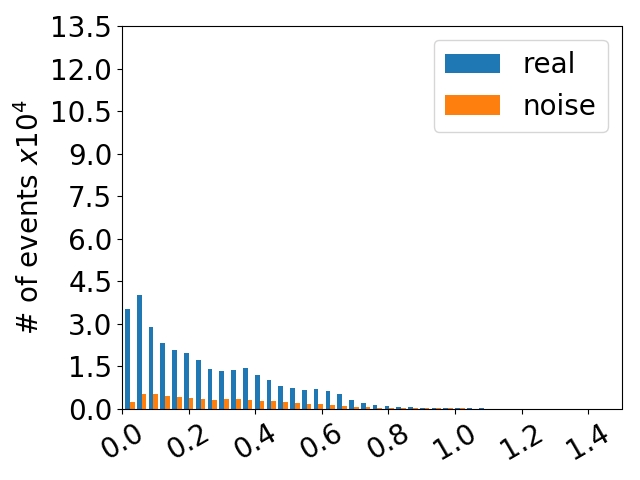}\\
        \vspace{0.02cm}
        \includegraphics[width=1.35in, height=0.8in]{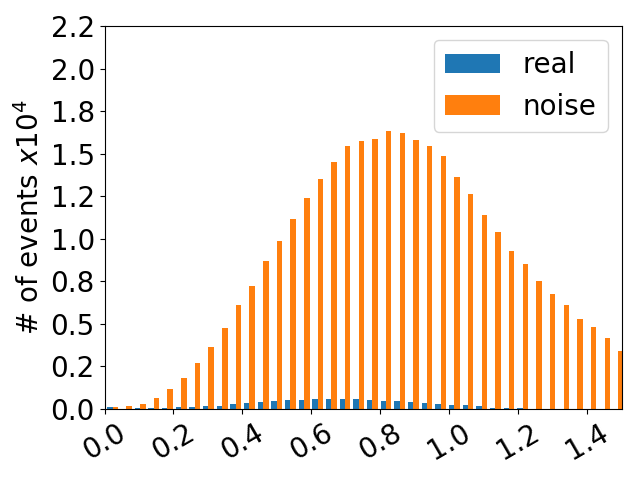}\\
        \vspace{0.02cm}
        \includegraphics[width=1.35in, height=0.8in]{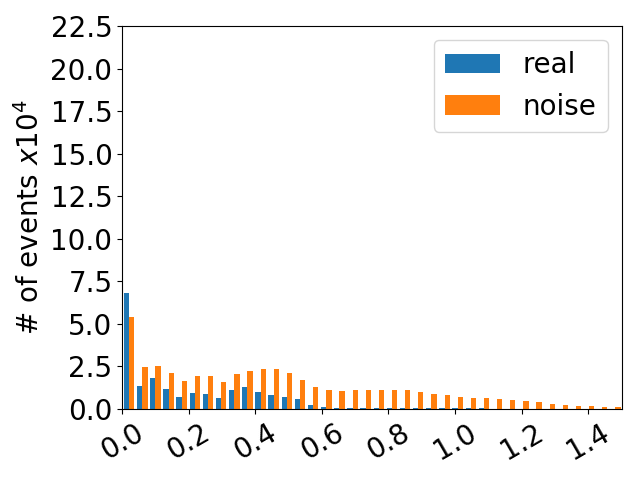}\\
        \vspace{0.02cm}
    \end{minipage}%
}%
\caption{The events distribution for different functions $f$ on four datasets. The x-axis is the value of $f(D_{n,n-1}, D_{n,n-2})$ for an event, smaller value suggesting closer spatial correlation between the event and the events in the past event window. The y-axis is the number of events. Each bar indicates the number of events with their $f(D_{n,n-1}, D_{n,n-2})$ value falling into the same range.}
\label{fig:distribution of bs1}
\end{figure}

\textbf{Observation} To choose the function for best separating the real events and noise events, we analyze the following distribution (Figure~\ref{fig:distribution of bs1}). The data are collected from bs1 on \textbf{four datasets} (described in section~\ref{sec:dataset}). Each row represents a dataset. From top to bottom, they are Gesture \cite{dataset}, paper subset from Roshambo \cite{bib:RoShamBo}, a noisy subset from Pendulum \citep{pendulum}, and a dataset recorded using Celex-IV by ourselves respectively. 
We only show the paper subset due to the page limitation but the rock and scissor subset show similar distribution as the paper subset. 
Bs1 gives a tag for each event, suggesting the event is real or noise. Here we use $X=2$ for analysis. We derive the four combinations from these data and plot their distribution. The parameters for weighted average are set to be 3 and 1. We use 5k$\times$70 events for each dataset.

We can see different distributions for different datasets in Figure~\ref{fig:distribution of bs1}, showing different percentage of real events and noise events derived from the bs1. The desired distribution is that the majority of real events are distributed near 0, while most of the noise events are distributed away from 0.
Such distribution suggests that the processed spatial distance between a real event and the events in the past event window, i.e., the value of $f(D_{n,n-1}, D_{n,n-2})$, is small, which demonstrates that the spatial correlation between the real event and the past events are very close. In contrast, such distribution shows that the processed spatial distance between a noise event and the events in the past event window is large, indicating no spatial correlation between the noise event and the past events.

Thus, the first column in the distribution is most important for separating real events and noise events. 

We introduce three rules for deciding which function is the best candidate. i) The real events in the first column should be the largest one among all distributions of real events. ii) Larger ratio between real events and noise events in the first column implies better distinguishing of real events and noise events. iii) When the ratio is similar among different modes, the mode with the highest real events in the first column is chosen as the best.

The first criterion is more important than the second because if the highest bar of real events is not the first column, to reach the distribution, more ranges are included and thus noise events increase much more rapidly than real events as the noise events have more distribution in larger ranges, which does more harm than good.


In summary, we decide to use minimum as the function $f$. Thus, the steps for SeqXFilter are outlined as follows (as shown in Figure~\ref{fig:overview}). For each event, i) calculate the spatial distance between the current event and the events in the past event window and choose the minimum one; ii) Check the minimum spatial distance with a certain spatial threshold $\sigma$. If the distance is less than $\sigma$, pass the event to the output, otherwise, discard it; iii) write the x and y coordinates of the new event in the oldest event location in the past event window.
The oldest event location is indexed by a counter.

\section{Experiments}
\subsection{Dataset}
\label{sec:dataset}
We use four datasets, a DVS dataset DvsGesture \cite{dataset}, a DAVIS240 dataset Roshambo \cite{bib:RoShamBo}, a DAVIS346 dataset Pendulum \citep{pendulum} \footnote{The pendulum data was collected at
the Telluride workshop}, and a dataset recorded from CeleX-IV, denoted as CRPS.

DvsGesture comprises 11 hand gesture categories from 29 subjects under 3 illumination conditions and the spatial dimension of the output is 128$\times$128. Roshambo is a dataset of rock, paper, scissors, and background images. We use three sub-recordings of rock, paper, and scissors. Pendulum is a small ball doing a pendulum movement, which is suspended from fixed support and swings freely back and forth under the influence of gravity. The noisy subset is recorded under dark illumination. CRPS is similar to Roshambo only the output is larger.

To make the event stream visible, it is common to generate a picture frame from the events, either of fixed time length or of a constant number of events. We choose to use the fixed number of events. Each event has a location that corresponds to a pixel. A pixel in the frame will be activated if there is an event with its location corresponding to the pixel. By meaning activates, the pixel value is set to be 255, otherwise, 0. After a bundle of events (fixed count) is processed, a frame is generated.

\subsection{Result}
\subsubsection{Visual Effect}
\begin{figure}
\centering
\subfigure[init]{
    \begin{minipage}[t]{0.19\textwidth}
        \centering
        \includegraphics[width=0.7in,height=0.7in]{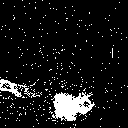}\\
        \vspace{0.02cm}
        \includegraphics[width=1.05in]{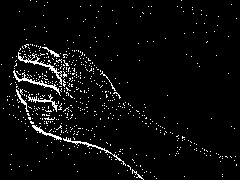}\\
        \vspace{0.02cm}
        \includegraphics[width=1.05in]{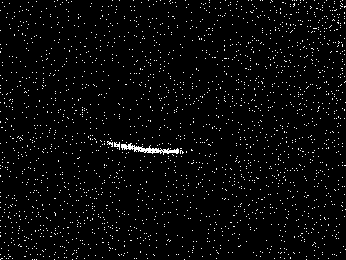}\\
        \vspace{0.02cm}
        \includegraphics[width=1.05in]{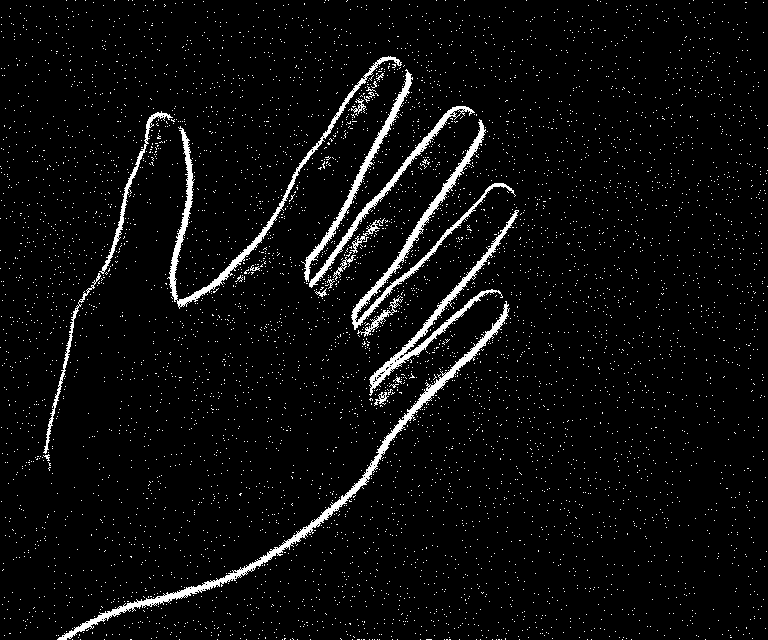}\\
        \vspace{0.02cm}
    \end{minipage}%
}%
\subfigure[bs1]{
    \begin{minipage}[t]{0.19\textwidth}
        \centering
        \includegraphics[width=0.7in,height=0.7in]{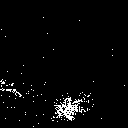}\\
        \vspace{0.02cm}
        \includegraphics[width=1.05in]{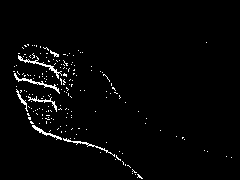}\\
        \vspace{0.02cm}
        \includegraphics[width=1.05in]{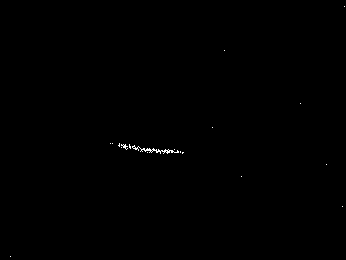}\\
        \vspace{0.02cm}
        \includegraphics[width=1.05in]{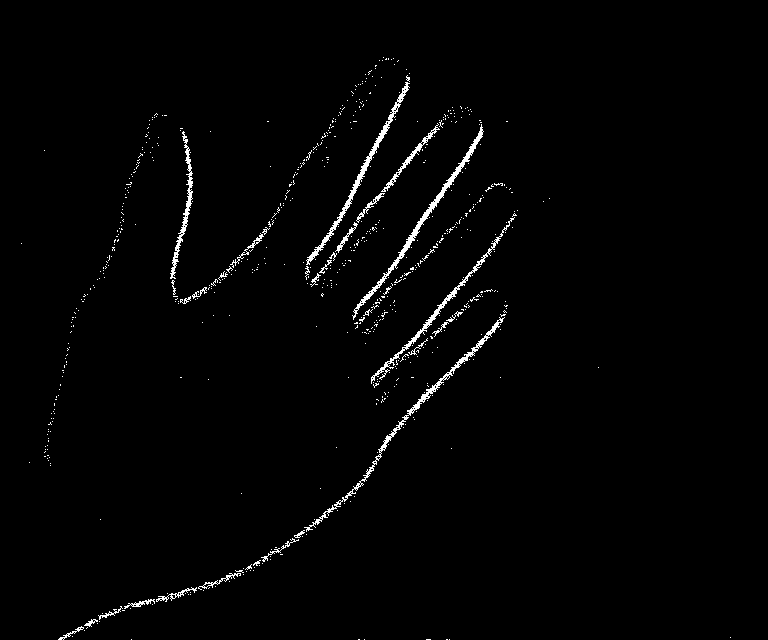}\\
        \vspace{0.02cm}
    \end{minipage}%
}%
\subfigure[bs2]{
    \begin{minipage}[t]{0.19\textwidth}
        \centering
        \includegraphics[width=0.7in,height=0.7in]{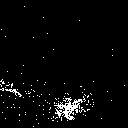}\\
        \vspace{0.02cm}
        \includegraphics[width=1.05in]{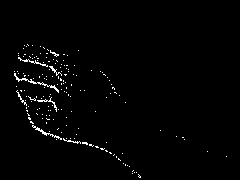}\\
        \vspace{0.02cm}
        \includegraphics[width=1.05in]{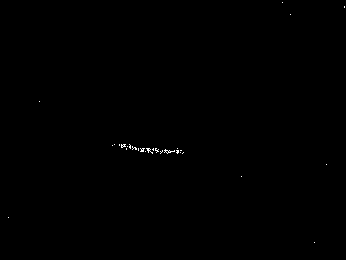}\\
        \vspace{0.02cm}
        \includegraphics[width=1.05in]{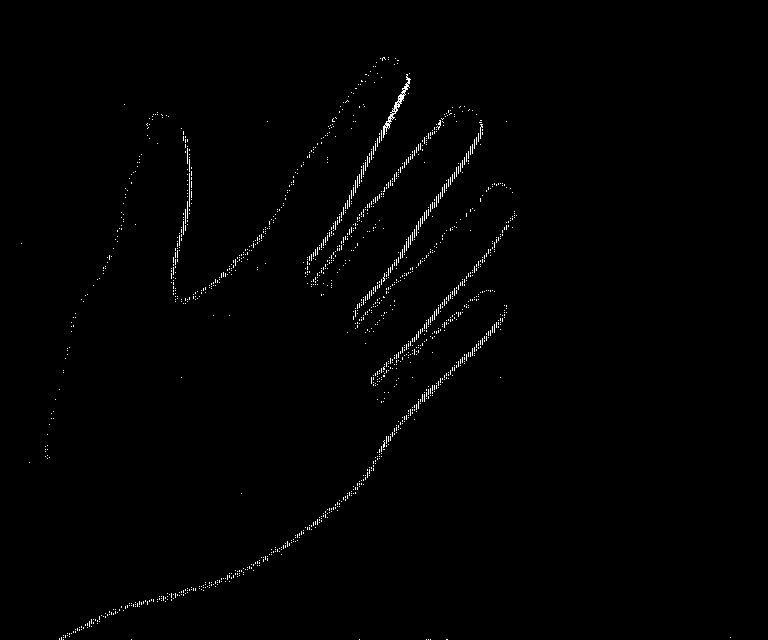}\\
        \vspace{0.02cm}
    \end{minipage}%
}%
\subfigure[seq $\sigma_{1}$]{
    \begin{minipage}[t]{0.19\textwidth}
        \centering
        \includegraphics[width=0.7in,height=0.7in]{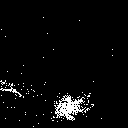}\\
        \vspace{0.02cm}
        \includegraphics[width=1.05in]{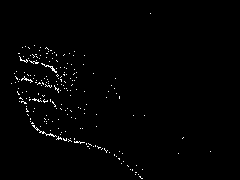}\\
        \vspace{0.02cm}
        \includegraphics[width=1.05in]{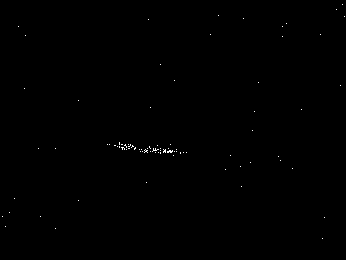}\\
        \vspace{0.02cm}
        \includegraphics[width=1.05in]{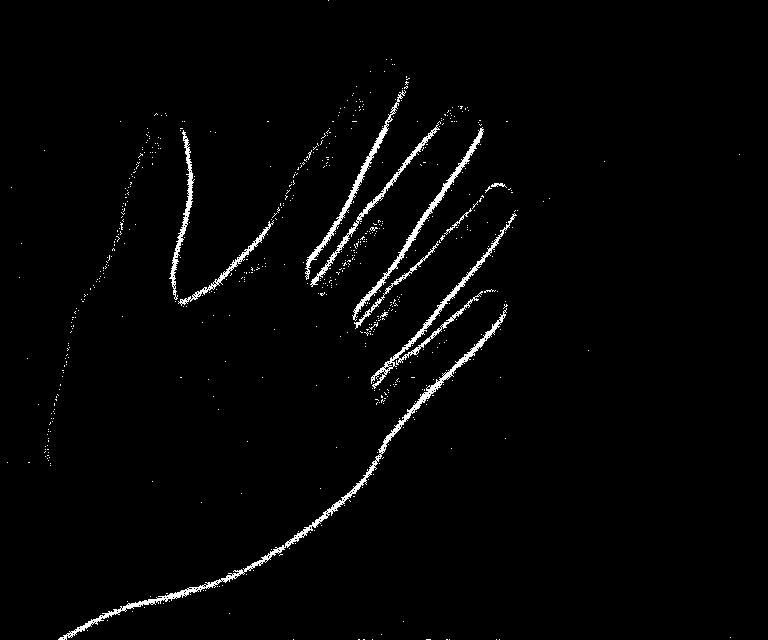}\\
        \vspace{0.02cm}
    \end{minipage}%
}%
\subfigure[seq $\sigma_{2}$]{
    \begin{minipage}[t]{0.19\textwidth}
        \centering
        \includegraphics[width=0.7in,height=0.7in]{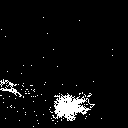}\\
        \vspace{0.02cm}
        \includegraphics[width=1.05in]{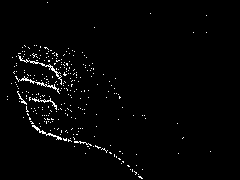}\\
        \vspace{0.02cm}
        \includegraphics[width=1.05in]{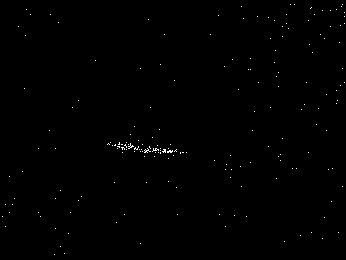}\\
        \vspace{0.02cm}
        \includegraphics[width=1.05in]{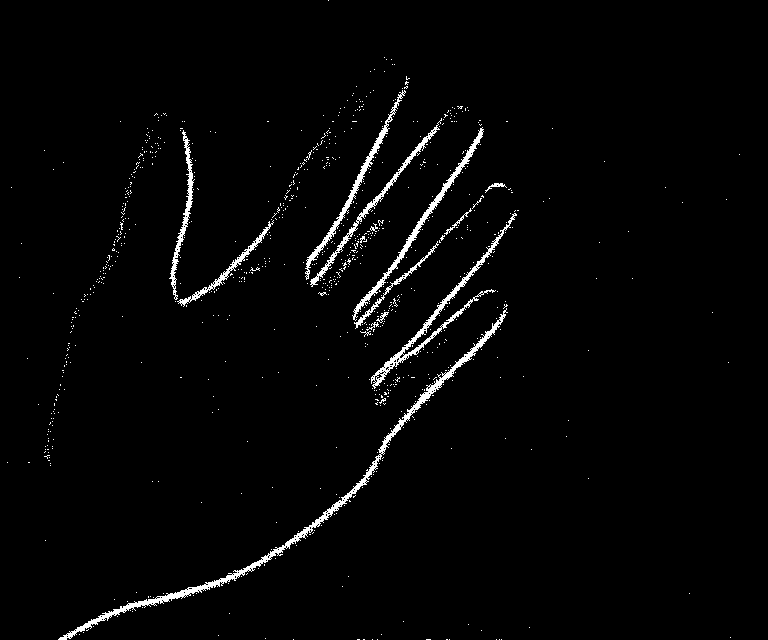}\\
        \vspace{0.02cm}
    \end{minipage}%
}%
\caption{The visual effect of denoising. The first column is the initial frame without denoising. The other columns are denoising effect from different filters. Each row represents a different dataset.}
\vspace{-0.2cm}
\label{fig:visualeffect}
\end{figure}

Figure~\ref{fig:visualeffect} shows the denoised frames for the three datasets using three filters. Due to page limitation, we only give one example for each dataset as shown in Figure~\ref{fig:visualeffect}. The filters are bs1, bs2, and SeqXFilter with $\sigma_{1}$ and $\sigma_{2}$. The time threshold for bs1 is 0.2ms (Gesture), 1ms (Roshambo and Pendulum), and 10ms (CRPS) respectively. The time threshold for bs2 is 2x of that for bs1. We can see that the visual effects are very similar, which shows that our filter is as effective as bs1 and bs2 but with drastically reduced memory cost and similar number of operations.

\subsubsection{Quantitative analysis}
Peak Signal-to-Noise Ratio (PSNR) and Structural Similarity (SSIM) \cite{wang2004image} index are calculated for evaluation. Since we can only get "clean" frames from baseline filters for calculating the two metrics, we adopt two baseline filters, bs1 and bs2 for demonstrating the validness of SeqXFilter. Thus, we calculate two PSNR values, denoted as PSNR1 and PSNR2. PSNR1 is calculated using bs1 frames and SeqXFilter frames, and PSNR2 is calculated using bs2 frames and SeqXFilter frames. The frames are generated using fixed number of events. There are three settings of the number, namely C1, C2, and C3, to demonstrate the effectiveness of our method. For datasets except CRPS, C1 is 1k, C2 is 3k, and C3 is 5k. For CRPS, they are 10x larger due to the large output of Celex.
We do the same processing for SSIM as PSNR. For PSNR and SSIM, the bigger the better.


\begin{table}[htbp]
  \centering
  \caption{PSNR (dB) and SSIM comparison for all datasets.}
  \small
  \begin{tabular}{llllllll}
  \toprule
    \multicolumn{1}{l}{\# of events} &       & gesture & paper & rock  & sci   & pendulum & CRPS \\
    \midrule
    \multirow{4}[0]{*}{C1} & PSNR1 & 24.20  & 20.32  & 20.18  & 20.55  & 34.69  & 27.15  \\
          & PSNR2 & 23.30  & 20.46  & 20.84  & 21.30  & 34.78  & 24.03  \\
          & SSIM1 & 0.95  & 0.90  & 0.89  & 0.91  & 1.00  & 0.97  \\
          & SSIM2 & 0.94  & 0.90  & 0.89  & 0.91  & 1.00  & 0.95  \\
          \midrule
    \multirow{4}[0]{*}{C2} & PSNR1 & 20.81  & 15.76  & 16.44  & 16.41  & 30.60  & 22.23  \\
          & PSNR2 & 19.92  & 16.06  & 16.92  & 16.76  & 30.74  & 18.68  \\
          & SSIM1 & 0.94  & 0.83  & 0.84  & 0.86  & 0.99  & 0.95  \\
          & SSIM2 & 0.91  & 0.83  & 0.83  & 0.85  & 0.99  & 0.90  \\
          \midrule
    \multirow{4}[0]{*}{C3} & PSNR1 & 19.80  & 13.93  & 14.81  & 14.61  & 28.43  & 19.79  \\
          & PSNR2 & 18.72  & 14.21  & 15.21  & 14.89  & 28.60  & 16.04  \\
          & SSIM1 & 0.93  & 0.79  & 0.82  & 0.83  & 0.98  & 0.93  \\
          & SSIM2 & 0.87  & 0.79  & 0.81  & 0.83  & 0.98  & 0.87  \\
          \bottomrule
    \end{tabular}%
\label{tab:psnrssimall}%
\end{table}%

Table~\ref{tab:psnrssimall} shows the PSNR and SSIM results for all datasets when SeqXFilter using $\sigma_{1}$ (0.005 for CRPS and 0.05 for others). C1, C2, and C3 are introduced above. As we can see from Table~\ref{tab:psnrssimall}, PSNR1 and PSNR2 are close to each other, which suggests that the performance of our filter is close to that of bs1 as well as bs2. 
Similar observations can be made on SSIM. We note that the SSIM value for pendulum is very high. The reason is that the whole frame is almost dark (the pixel value is 0) and the calculation rule of SSIM. SSIM models distortion as a combination of three factors, i.e., brightness, contrast, and structure, and use the mean as an estimate of brightness, the standard deviation as an estimate of contrast, and the co-variance as a measure of structural similarity. However, only a small fraction of pixels in the pendulum case, especially for C1 frames, are active, i.e., 255, and all other pixels are 0. Thus, the mean, standard deviation, and co-variance of bs1 frame are easily close to that of SeqXFilter frame since they are all close to 0.

The PSNR decrease when the number of events constituting a frame increase, especially for Roshambo dataset. The reason will be discussed in section~\ref{sec:sigma}.

\subsection{the effect of $\sigma$}
\label{sec:sigma}
\begin{wrapfigure}{l}{0.35\textwidth}
\centering
        \includegraphics[width=0.34\textwidth]{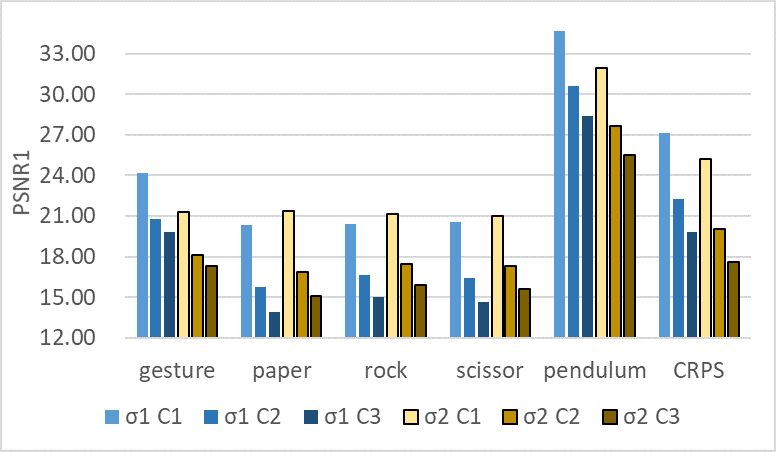}
    \caption{PSNR1 comparison using different $\sigma$. The x-axis is dataset name, and the y-axis is the PSNR1 value. C1, C2, and C3 are the number of events building up a frame.}
    \label{fig:psnr1sigma}
\vspace{-0.02in}
\end{wrapfigure}
Figure~\ref{fig:psnr1sigma} shows the comparison of PSNR1 using different $\sigma$. As experiments above, the number of events forming a frame (C1, C2, and C3) for CRPS is 10k, 30k, and 50k instead of 1k, 3k, and 5k for other datasets. The $\sigma_{1}$ and $\sigma_{2}$ for CRPS are 0.005 and 0.01 instead of 0.05 and 0.1 for other datasets.

For Gesture, pendulum, and CRPS, increase of $\sigma$ leads to decrease of PSNR1, while for subsets from Roshambo, the rise of $\sigma$ brings the ascent of PSNR1. i) For Gesture, this is because the time threshold for baseline filters are small and many real events are also filtered leading to less bright pixels caused by real events as shown in Figure~\ref{fig:visualeffect}. However, the bright pixels in the frames from SeqXFilter are more than that in baseline filters but still are centered in the moving object. The reason is similar for CRPS. ii) For pendulum, increase of $\sigma$ leads to decrease of PSNR because it passes more noise events. And more noise pixels are activated in the SeqXFilter frame than in the bs1 frame. iii) For three Roshambo subsets, paper, rock, and scissor, the time threshold for baseline filters are 1ms, the baseline filters generate frames with clear outline of object, the SeqXFilter filter with $\sigma_{1}$ of 0.05 generates less clear outline, and the SeqXFilter filter with $\sigma_{2}$ of 0.1 generates clearer outline that close to bs1 frame. Thus, the PSNR1 is increased.


In conclusion, the PSNR shows the upward or downward trend due to the specific time threshold chosen for bs1 and the $\sigma$ chosen for SeqXFilter. But all demonstrate the similar performance of our filter and the bs1. Similar performance are also found for bs2. Moreover, for noisy datasets, large threshold $\sigma$ does more harm than good, incurring more noise than real events. For less noisy and clean datasets, the range of $\sigma$ is more flexible.

\subsection{the effect of window length}
\begin{wrapfigure}{l}{0.35\textwidth}
\centering
        \includegraphics[width=0.34\textwidth]{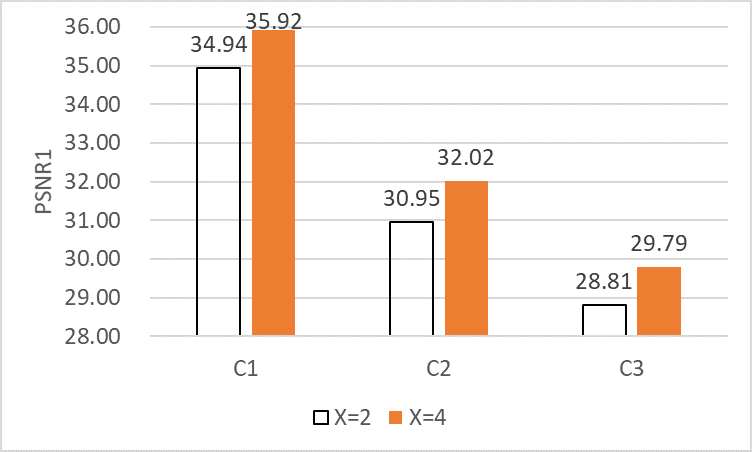}
    \caption{PSNR1 comparison using different $X$ on pendulum.}
    \label{fig:psnr1window}
\vspace{-0.02in}
\end{wrapfigure}
For extremely noisy dataset like pendulum, more than one noise events likely occur between two real events. Thus, increasing the length of the past event window is believed to lead to better denoising performance in such case. In the above experiments, we set the window length $X$ to be 2. In this section, we set it to be 4 and 2 respectively and explore the performance on the pendulum dataset.
The $\sigma$ for $X=4$ is 0.02, and for $X=2$ is 0.04. This setting is because when the window is enlarged, for extremely noisy case, a real event is likely to get a more spatial related event in the window of 4 than in the window of 2. Moreover, a noise event is also likely to be supported by a past event if the $\sigma$ is the same as that in the window of 2, since most of the events are noise events.

\begin{figure}
\centering
\subfigure[X=2]{
\label{fig:vewindow2}
    \begin{minipage}[t]{0.49\textwidth}
        \centering
        \includegraphics[width=2.65in, height=1.6in]{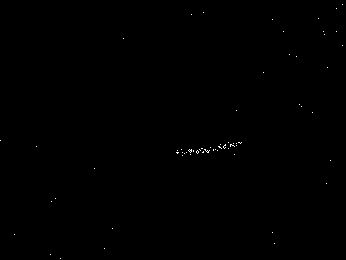}\\
        \vspace{0.02cm}
    \end{minipage}%
}%
\subfigure[X=4]{
\label{fig:vewindow4}
    \begin{minipage}[t]{0.49\textwidth}
        \centering
        \includegraphics[width=2.65in, height=1.6in]{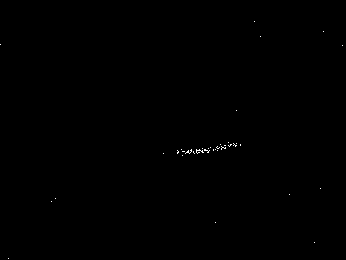}\\
        \vspace{0.02cm}
    \end{minipage}%
}%
\caption{The visual effect of denoising with different window length $X$.}
\vspace{-0.2cm}
\label{fig:vewindow}
\end{figure}

As shown in Figure~\ref{fig:psnr1window}, the difference for different event count is all about 1dB. As shown in Figure~\ref{fig:vewindow2}, window $d=2$ has obvious noise pixels in the upper right of the frame. With the bigger window $d=4$, there are less noisy pixels as shown in Figure~\ref{fig:vewindow4}, while the moving objects in both scenes are almost bright. As mentioned above, the real events still get support with smaller $\sigma$ that helps to reduce the Wrong judgment of noise events. The bigger $\sigma$ for window of 2 is able to maintain real events but also passes more noise events.

\subsection{Advantage}
We utilize the spatio-correlation from a naive perspective by storing the location of the past few events before the current event to support the spatio-temporal correlation check. Our goal is to achieve similar performance as background filters since our method accounts for negligible memory cost. The past event window stores the x and y coordinates of the $X$ past events. The bit width for the x and y coordinates is 16 bits. So the storage for $X$ past events is 32$\cdot$X bits. As $X=2$ is satisfactory for most occasions and $X=4$ is proved to improve the performance for difficult occasions (the noisy pendulum), basically, the storage equals to only 2 timestamps or 4 timestamps. For these different DVS with different output size, $X=2$ achieves similar performance as baseline filters, which is actually an O(1) space complexity method.

For computing cost, for $X=2$, it requires 6 additions, 2 divisions, 2 comparisons, and 2 writes. The counter updating requires an addition and a complementation. Divisions are due to the normalization of distance. This can be eliminated by adjusting the threshold.

\section{Conclusion}
  Neuromorphic event-based dynamic vision sensors (DVS) have much faster sampling rates and a higher dynamic range than frame-based imaging sensors. However, they are sensitive to background activity (BA) events that are unwanted. Previous spatio-temporal filters are either memory-intensive or computing-intensive. We propose \emph{SeqXFilter}, which is a spatio-temporal correlation filter with only a past event window. It has an O(1) space complexity and simple computations. We explore the spatial correlation of an event with its past few events by analyzing the distribution of the events when applying different functions on the spatial distances. We find the best function to check the spatio-temporal correlation for an event for \emph{SeqXFilter}, best separating real events and noise events. We not only give the visual denoising effect of the filter but also use two metrics for quantitatively analyzing the filter's performance. Four neuromorphic event-based datasets, recorded from four DVS with different output sizes, are used for validation of our method. The experimental results show that \emph{SeqXFilter} achieves similar performance as baseline NNb filters, but with extremely small memory cost and simple computation logic.

\small
\bibliographystyle{unsrt}
\bibliography{filter-base}




\end{document}